\documentstyle[aclap,psfig,graphicx]{article}

\title{\vspace{-0.5in}Finite State Transducers \\
Approximating Hidden Markov Models}
\author{Andr\'e Kempe \\
Rank Xerox Research Centre -- Grenoble Laboratory \\
6, chemin de Maupertuis -- 38240 Meylan -- France \\
{\normalsize {\tt andre.kempe@grenoble.rxrc.xerox.com}} \\
{\normalsize {\tt http://www.rxrc.xerox.com/research/mltt}} }

\begin{document}

\newcommand{\hide}[1]{$\langle${\sc Hidden~Name}$\rangle$}

\newcommand{\mycomment}[1]
{ \rule{0mm}{0mm} \\
\begin{center}
\fbox{
\begin{minipage}{120mm}
\noindent \underline{\large \bf \it COMMENT :} \\

\vspace{-2mm}
 #1 
\end{minipage} }
\end{center}
 \rule{0mm}{0mm} \\ }

\newcounter{NumRE}     

\newcommand{\REa}[2]           
{ \hfill
\refstepcounter{NumRE}
\makebox[120mm][#1]{{\tt #2}}  
\makebox[15mm][r]{[\theNumRE]} \\ } 

\newcommand{\REaL}[3]          
{ \REa{#1}{#2} \label{#3} }    

\newcommand{\REb}[1]           
{ \hfill
\refstepcounter{NumRE}
\parbox[t]{120mm}{{\tt #1}}                 
\parbox[t]{15mm}{\hfill [\theNumRE]} \\ }   

\newcommand{\REbL}[2]          
{ \REb{#1} \label{#2} }

\newcommand{\EQb}[1]           
{ \hfill
\refstepcounter{NumRE}
\parbox[t]{120mm}{{${#1}$}}                 
\parbox[t]{15mm}{\hfill [\theNumRE]} \\ }

\newcommand{\eqcom}[2]         
{\hfill \begin{minipage}{#1} {\it #2} \end{minipage} }

\newcommand{\RStep}[5]         
{\noindent (#1) #2 \\          
\vspace{-3mm} \\

\REbL{#4}{#3}                  
\vspace{-3mm} \\

\eqcom{140mm}{#5} }

\newcommand{\otherfootnote}[1]  
{$^{\ref{#1}}$}

\newcommand{\LeftBr}[1]{ {\tt \LARGE \bf <}$_{#1}$}
\newcommand{\RightBr}[1]{ {\tt \LARGE \bf >}$_{#1}$}

\def\contains{\$}               
\def\xnot{\tilde{\enspace}}
\def\notcont{\xnot \contains}

\def\ctxU{$||$}
\def\ctxR{$//$}
\def\ctxL{$\backslash \backslash$~}
\def\ctxD{$\backslash /$~}

\def\igin{\tt .\rule{-0.8mm}{0mm}/\rule{-1.6mm}{0mm}.}

\def\eps{\epsilon}
\def\bs{\backslash}
\def\SmSpc{\rule{1.5mm}{0mm}}
\def\cct{$^{\frown}$}
\def\CtxBd{{\tt.\#.}}

\def\kleenestar{\!\!*}
\def\kleeneplus{\!{\tt+}}
\def\crosspr{\; {\tt .x.}\; }
\def\compose{\; {\tt .o.}\; }

\def\repl{\; {\tt -\!\!\!>}\; }
\def\invrepl{\; {\tt <\!\!\!-}\; }
\def\birepl{\; {\tt <\!\!-\!\!\!>}\; }
\def\symp{\!:\!}
\newcommand{\OLPair}[2]{ \langle #1,#2 \rangle }
\def\OneL{.{\it1L}}
\def\TwoL{.{\it2L}}
\newcommand{\Xlevel}[1]{\overline{#1}}

\def\longbar{\rule[1mm]{9mm}{0.2mm}}

\newcommand{\lxu}[1]{\rule{-2mm}{0mm}\enspace^{#1}\rule{-0.5mm}{0mm}} 
\newcommand{\lxl}[1]{\rule{-2mm}{0mm}\enspace_{#1}\rule{-0.5mm}{0mm}} 
\newcommand{\lxul}[2]{\rule{-2mm}{0mm}\enspace_{#2}^{#1}\rule{-0.5mm}{0mm}}

\newcommand{\spc}[1]{\rule{#1}{0mm}}

\def\SmSpc{\spc{1.5mm}}
\def\spA{\spc{3mm}}
\def\spB{\spc{6mm}}

\def\SpcUp{\rule{0mm}{4.5mm}}
\def\SpcDown{\rule[-1mm]{0mm}{2mm}}

\def\mybibbegin{\begin{minipage}{2mm} \end{minipage} \hfill
                \begin{minipage}{75mm}}

\def\mybibend{\end{minipage}}

\def\mybibitem{\vspace{1.8mm}
               \noindent \rule{-3.5mm}{0mm}}

\bibliographystyle{fullname}
\maketitle
\vspace{-0.5in}
\begin{abstract}
This paper describes the conversion of a Hidden Markov Model
into a sequential transducer
that closely approximates the behavior of the stochastic model.
This transformation is especially advantageous for 
part-of-speech tagging
because the resulting transducer can be composed with 
other transducers that encode correction rules
for the most frequent tagging errors.
The speed of tagging is also improved.
The described methods have been implemented and
successfully tested on six languages.
\end{abstract}

\vspace{2mm}
\section{Introduction \label{s-intro}}
\vspace{1mm}

Finite-state automata have been successfully applied in many areas of
computational linguistics.

This paper describes two algorithms\footnote{
There is a different (unpublished) algorithm by
\mbox{Julian} M. Kupiec and John T. Maxwell (p.c.).
} which approximate a {\it Hidden Markov Model} (HMM) used for
part-of-speech tagging by a {\it finite-state transducer} (FST).
These algorithms may be useful beyond the current
description on any kind of analysis of written or spoken language
based on both finite-state technology and HMMs,
such as corpus analysis, speech recognition, etc.
Both algorithms have been fully implemented.

An HMM used for tagging encodes, like a transducer,
a relation between two languages.
One language contains sequences of ambiguity classes
obtained by looking up in a lexicon all words of a sentence.
The other language contains sequences of tags
obtained by statistically disambiguating the class sequences.
From the outside, an HMM tagger behaves like a sequential transducer
that deterministically maps every class sequence to a tag sequence,
e.g.:
\begin{equation}
\frac{\tt{[DET,PRO] ~[ADJ,NOUN] ~[ADJ,NOUN] ~...... ~[END]}}
 {\tt{\spc{5mm}DET\spc{12mm}ADJ\spc{12mm}NOUN\spc{3mm} ~......\spc{3mm}END}}
\end{equation}

The aim of the conversion is not to generate FSTs that behave
in the same way, or in as similar a way as possible like HMMs,
but rather FSTs that perform tagging in as accurate a way as possible.
The motivation to derive these FSTs from HMMs is that HMMs
can be trained and converted with little manual effort.

The tagging speed when using transducers is up to five times higher
than when using the underlying HMMs.
The main advantage of  transforming an HMM is that the resulting
transducer can be handled by finite state calculus.
Among others, it can be composed with transducers that encode:
\begin{itemize}
\item
correction rules for the most frequent tagging errors
which are automatically generated
(Brill, 1992; Roche and Schabes, 1995)
or manually written (Chanod and Tapanainen, 1995),
in order to significantly improve tagging accuracy\footnote{
Automatically derived rules require less work than manually written ones
but are unlikely to yield better results because they would consider
relatively limited context and simple relations only.
}.
These rules may include
long-distance dependencies not handled by HMM taggers,
and can conveniently be expressed by the replace operator
(Kaplan and Kay, 1994; Karttunen, 1995; Kempe and Karttunen, 1996).
\item
further steps of text analysis, e.g. light parsing or extraction
of noun phrases or other phrases (A\"{\i}t-Mokhtar and Chanod, 1997).
\end{itemize}

These compositions enable complex text analysis to be performed
by a single transducer.

An HMM transducer builds on the data (probability matrices)
of the underlying HMM.
The accuracy of this data
has an impact on the tagging accuracy of both the HMM itself
and the derived transducer.
The training of the HMM can be done
on either a tagged or untagged corpus, and is not a topic
of this paper since it is exhaustively described in the literature
(Bahl and Mercer, 1976; Church, 1988).

An HMM can be identically represented by a weighted FST
in a straightforward way.
We are, however, interested in non-weighted transducers.

\vspace{2mm}
\section{n-Type Approximation \label{s-ntype}}
\vspace{1mm}

This section presents a method that approximates a (1st order) HMM
by a transducer, called {\it n-type} approximation\footnote{
Name given by the author.}.

Like in an HMM, we take into account
initial probabilities $\pi$, transition probabilities $a$
and class (i.e. observation symbol) probabilities $b$.
We do, however, not estimate probabilities over paths.
The tag of the first word is selected based on its initial
and class probability.
The next tag is selected on its transition probability
given the first tag, and its class probability, etc.
Unlike in an HMM, once a decision on a tag has been made,
it influences the following decisions
but is itself irreversible.

\begin{figure*}[htbp]
\begin{center}
\includegraphics[scale=0.6,angle=0]{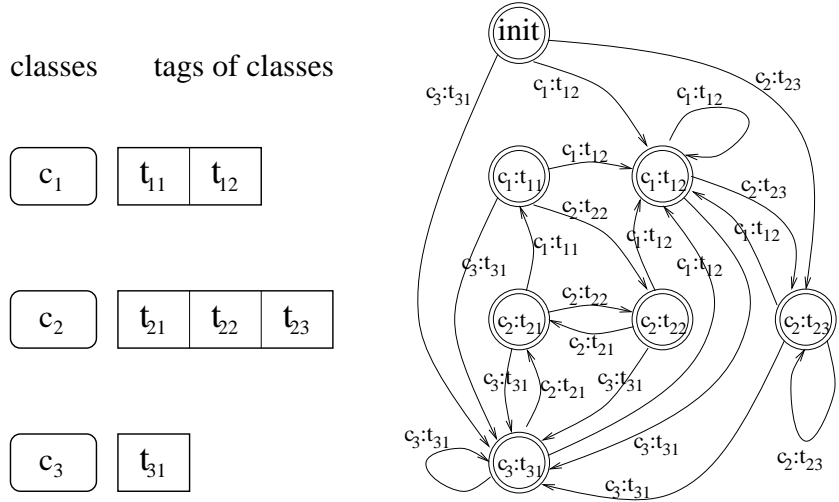}
\caption{Generation of an n1-type transducer \label{f-nfst}}
\end{center}
\end{figure*}

A transducer encoding this behaviour can be generated
as sketched in figure \ref{f-nfst}.
In this example we have a set of three classes,
$c_1$ with the two tags $t_{11}$ and $t_{12}$,
$c_2$ with the three tags $t_{21}$, $t_{22}$ and $t_{23}$, and
$c_3$ with one tag $t_{31}$.
Different classes may contain the same tag,
e.g. $t_{12}$ and $t_{23}$ may refer to the same tag.

For every possible pair of a class and a tag
(e.g. $c_1\symp t_{12}$ or {\tt [ADJ,NOUN]:NOUN})
a state is created and labelled with this same pair
(fig.~\ref{f-nfst}).
An initial state which does not correspond with any pair,
is also created.
All states are final, marked by double circles.

For every state, as many outgoing arcs are created
as there are classes (three in fig.~\ref{f-nfst}).
Each such arc for a particular class points to the most probable
pair of this same class.
If the arc comes from the initial state,
the most probable pair of a class and a tag
(destination state) is estimated by:
\begin{equation}
\arg \max_k p_1 (c_i , t_{ik}) = \pi (t_{ik}) \; b (c_i | t_{ik})
  \label{e-n1-init}
\end{equation}

\noindent
If the arc comes from a state other than the initial state,
the most probable pair is estimated by:
\begin{equation}
\arg \max_k p_2 (c_i , t_{ik}) = a (t_{ik} | t_{previous}) \; b (c_i | t_{ik})
  \label{e-n1-step}
\end{equation}

In the example (fig.~\ref{f-nfst})
$c_1\symp t_{12}$ is the most likely pair of class $c_1$, and
$c_2\symp t_{23}$ the most likely pair of class $c_2$
when coming from the initial state, and
$c_2\symp t_{21}$ the most likely pair of class $c_2$
when coming from the state of $c_3\symp t_{31}$.

Every arc is labelled with the same symbol pair
as its destination state, with the class symbol in the upper language
and the tag symbol in the lower language.
E.g. every arc leading to the state of $c_1\symp t_{12}$
is labelled with {\tt c$_1$:t$_{12}$}.

Finally, all state labels can be deleted since the behaviour
described above is encoded in the arc labels and the network
structure.
The network can be minimized and determinized. \\

We call the model an {\it n1-type model}, the resulting FST
an {\it n1-type transducer} and the algorithm leading
from the HMM to this transducer, an {\it n1-type approximation}
of a 1st order HMM.

Adapted to a 2nd order HMM, this algorithm
would give an {\it n2-type approximation}.
Adapted to a zero order HMM, which means only to use
class probabilities $b$,
the algorithm would give an {\it n0-type approximation}.

n-Type transducers have deterministic states only.

\vspace{3mm}
\section{s-Type Approximation \label{s-stype}}
\vspace{1mm}

This section presents a method that approximates an HMM
by a transducer, called {\it s-type} approximation\footnote{
Name given by the author.}.

Tagging a sentence based on a 1st order HMM
includes finding the most probable tag sequence $T$
given the class sequence $C$ of the sentence.
The joint probability of $C$ and $T$ can be estimated by:
\begin{eqnarray}
p (C,T) = p (c_1 .... c_n , t_1 .... t_n) = \spc{15mm} \nonumber \\
   \spc{10mm}  \pi (t_1) \; b (c_1|t_1) \cdot
         \prod\limits_{i=2}^{n} a (t_i|t_{i-1}) \; b (c_i|t_i)
\end{eqnarray}

The decision on a tag of a particular word cannot be made
separately from the other tags.
Tags can influence each other over a long distance via
transition probabilities.
Often, however, it is unnecessary to decide on the tags
of the whole sentence at once.
In the case of a 1st order HMM,
unambiguous classes (containing one tag only),
plus the sentence beginning and end positions,
constitute barriers to the propagation of HMM probabilities.
Two tags with one or more barriers inbetween do not influence
each other's probability.

\vspace{1mm}
\subsection{s-Type Sentence Model \label{s-sentmodel}}
\vspace{1mm}

To tag a sentence, one can split its class sequence at the barriers
into subsequences, then tag them separately and concatenate them
again.
The result is equivalent to the one obtained by tagging the
sentence as a whole.

We distinguish between initial and middle subsequences.
The final subsequence of a sentence is equivalent to
a middle one, if we assume that the sentence end symbol (.~or !~or~?)
always corresponds to an unambiguous class $c_u$.
This allows us to ignore the meaning of the sentence end position
as an HMM barrier because this role is taken by the
unambiguous class $c_u$ at the sentence end.

An initial subsequence $C_i$
starts with the sentence initial position,
has any number (incl. zero) of ambiguous classes $c_a$
and ends with the first unambiguous class $c_u$ of the sentence.
It can be described by the regular expression\footnote{
Regular expression operators used in this section
are explained in the annex. \label{n-regex}}:
\begin{equation}
C_i = c_a\kleenestar c_u
 \label{e-init-cseq}
\end{equation}

The joint probability of an initial class subsequence $C_i$ of length $r$,
together with an initial tag subsequence $T_i$, can be estimated by:
\begin{equation}
p (C_i, T_i) =
    \pi (t_1) \; b (c_1 | t_1) \cdot
         \prod\limits_{j=2}^{r} a (t_j | t_{j-1}) \; b (c_j | t_j)
\end{equation}

A middle subsequence $C_m$
starts immediately after an unambiguous class $c_u$,
has any number (incl. zero) of ambiguous classes $c_a$
and ends with the following unambiguous class $c_u$:
\begin{equation}
C_m = c_a\kleenestar c_u
 \label{e-mid-cseq}
\end{equation}

\noindent
For correct probability estimation
we have to include the immediately preceding unambiguous class $c_u$,
actually belonging to the preceding subsequence $C_i$ or $C_m$.
We thereby obtain an extended middle subsequence\otherfootnote{n-regex}:
\begin{equation}
C_m^e = c_u^e \; c_a\kleenestar c_u
 \label{e-emid-cseq}
\end{equation}

The joint probability of an extended middle class subsequence
$C_m^e$ of length $s$, together with a tag subsequence $T_m^e$,
can be estimated by:
\begin{equation}
p (C_m^e, T_m^e) =
    b (c_1 | t_1) \cdot
       \prod\limits_{j=2}^{s} a (t_j | t_{j-1}) \; b (c_j | t_j)
\end{equation}

\vspace{2mm}
\subsection{Construction of an s-Type Transducer \label{s-make-stype}}
\vspace{1mm}

To build an s-type transducer, a large number
of initial class subsequences $C_i$ and extended middle
class subsequences $C_m^e$ are generated
in one of the following two ways: \\

\noindent
{\bf (a) Extraction from a corpus}

Based on a lexicon and a guesser, we annotate an untagged training
corpus with class labels.
From every sentence, we extract the initial class subsequence $C_i$
that ends with the first unambiguous class $c_u$ (eq. \ref{e-init-cseq}),
and all extended middle subsequences $C_m^e$ ranging
from any unambiguous class $c_u$ (in the sentence)
to the following unambiguous class (eq. \ref{e-emid-cseq}).

A frequency constraint (threshold) may be imposed on
the subsequence selection, so that the only subsequences retained
are those that occur at least a certain number of times in
the training corpus\footnote{
The frequency constraint may prevent the encoding of rare
subsequences which would encrease the size of the transducer
without contributing much to the tagging accuracy.
}. \\

\noindent
{\bf (b) Generation of possible subsequences}

Based on the set of classes,
we generate all possible initial and extended middle
class subsequences, $C_i$ and $C_m^e$
(eq. \ref{e-init-cseq}, \ref{e-emid-cseq})
up to a defined length. \\

Every class subsequence $C_i$ or $C_m^e$
is first disambiguated based on a 1st order HMM,
using the Viterbi algorithm (Viterbi, 1967; Rabiner, 1990) for efficiency,
and then linked
to its most probable tag subsequence $T_i$ or $T_m^e$
by means of the cross product operation\otherfootnote{n-regex}:
\begin{equation}
S_i = C_i \crosspr T_i = c_1\symp t_1 \; c_2\symp t_2 \;
  ...... \; c_n\symp t_n
 \label{e-init-Sseq}
\end{equation}
\begin{equation}
S_m^e = C_m^e \crosspr T_m^e = c_1^e\symp t_1^e \; c_2\symp t_2 \;
  ...... \; c_n\symp t_n
 \label{e-emid-Sseq}
\end{equation}

\noindent
In all extended middle subsequences $S_m^e$, e.g.:
\begin{equation}
S_m^e = \frac{C_m^e}{T_m^e} =  \spc{40mm} \enspace
\label{exm-mark-1}
\end{equation}
\vspace{-5mm}
\begin{eqnarray}
\enspace \spc{10mm}
\frac{\tt{[DET] ~[ADJ,NOUN] ~[ADJ,NOUN] ~[NOUN]}}
   {\tt{DET\spc{9mm}ADJ\spc{12mm}ADJ\spc{9mm}NOUN}}   \nonumber
\end{eqnarray}

\noindent
the first class symbol on the upper side
and the first tag symbol on the lower side, will be marked as
an extension that does not really belong to the middle sequence
but which is necessary to disambiguate it correctly.
Example (\ref{exm-mark-1}) becomes:
\begin{equation}
S_m^0 = \frac{C_m^0}{T_m^0} =  \spc{40mm} \enspace
\label{exm-mark-2}
\end{equation}
\vspace{-5mm}
\begin{eqnarray}
\enspace \spc{10mm}
\frac{\tt{0.[DET] ~[ADJ,NOUN] ~[ADJ,NOUN] ~[NOUN]}}
   {\tt{0.DET\spc{9mm}ADJ\spc{12mm}ADJ\spc{9mm}NOUN}}   \nonumber
\end{eqnarray}

\noindent
We then build the union $\lxu{\cup}S_i$ of all initial subsequences $S_i$
and the union $\lxu{\cup}S_m^e$ of all extended middle subsequences $S_m^e$,
and formulate a preliminary sentence model:
\begin{equation}
\lxu{\cup}S^0 \; = \; \lxu{\cup}S_i \;\; \lxu{\cup}S_m^0\kleenestar
 \label{e-pre-sent-model}
\end{equation}

\noindent
in which all middle subsequences $S_m^0$ are still marked and
extended in the sense that all occurrences of
all unambiguous classes are mentioned twice:
Once unmarked as $c_u$ at the end of every sequence $C_i$ or $C_m^0$,
and the second time marked as $c_u^0$ at the beginning of every
following sequence $C_m^0$.
The upper side of the sentence model $\lxu{\cup}S^0$
describes the complete (but extended) class sequences
of possible sentences, and
the lower side of $\lxu{\cup}S^0$ describes the corresponding
(extended) tag sequences.

To ensure a correct concatenation of initial and middle subsequences,
we formulate a concatenation constraint for the classes:
\begin{equation}
R_c = \bigcap\limits_j \; [ \notcont[\; \bs{c_u} \; c_u^0\;] \;]_j
\end{equation}

\noindent
stating that every middle subsequence must begin with
the same marked unambiguous class $c_u^0$
(e.g. {\tt 0\spc{-0.7mm}.\spc{-1mm}[DET]})
which occurs unmarked as $c_u$ (e.g. {\tt [DET]})
at the end of the preceding subsequence
since both symbols refer to the same occurrence
of this unambiguous class.

Having ensured correct concatenation,
we delete all marked classes on the upper side of the relation
by means of
\begin{equation}
D_c = [\;] \invrepl \left[\; \bigcup\limits_j \; [c_u^0]_j \;\right]
\end{equation}

\noindent
and all marked tags on the lower side by means of
\begin{equation}
D_t = \left[\; \bigcup\limits_j \; [t^0]_j \;\right] \repl [\;]
\end{equation}

\noindent
By composing the above relations with the preliminary
sentence model, we obtain the final sentence
model\otherfootnote{n-regex}:
\begin{equation}
S = D_c \compose R_c \compose \lxu{\cup}S^0 \compose D_t
 \label{e-fin-sent-model}
\end{equation}

We call the model an {\it s-type model}\/,
the corresponding FST an {\it s-type transducer}\/,
and the whole algorithm leading from the HMM to the transducer,
an {\it s-type approximation} of an HMM.

The s-type transducer tags any corpus which contains only known
subsequences, in exactly the same way, i.e. with the same errors,
as the corresponding HMM tagger does.
However, since an s-type transducer is incomplete,
it cannot tag sentences with one or more class subsequences
not contained in the union of the initial or middle subsequences.

\vspace{2mm}
\subsection{Completion of an s-Type Transducer \label{s-compl-stype}}
\vspace{1mm}

An incomplete s-type transducer $S$ can be completed with
subsequences from an auxiliary, complete n-type transducer $N$
as follows:

First, we extract the union of initial and
the union of extended middle subsequences,
$\lxul{\cup}{s}S_i$ and $\lxul{\cup}{s}S_m^e$
from the primary s-type transducer $S$, and the unions
$\lxul{\cup}{n}S_i$ and $\lxul{\cup}{n}S_m^e$
from the auxiliary n-type transducer $N$.
To extract the union $\lxu{\cup}S_i$ of initial subsequences
we use the following filter:
\begin{equation}
F_{S_i} = [\; \bs \OLPair{c_u}{t} \;] \kleenestar
      \;\; \OLPair{c_u}{t} \;\; [\; ? \symp [\;] \;] \;\kleenestar
 \label{e-FSi}
\end{equation}

\noindent
where $\OLPair{c_u}{t}$ is the 1-level format\footnote{
1-Level and 2-level format are explained in the annex.\label{n-formats}}
of the symbol pair $c_u\symp t$.
The extraction takes place by
\begin{equation}
\lxu{\cup}S_i = [\; N\OneL \compose F_{S_i} \;].l\TwoL
\end{equation}

\noindent
where the transducer $N$ is first converted into
1-level format\otherfootnote{n-formats},
then composed with the filter $F_{S_i}$ (eq. \ref{e-FSi}).
We extract the lower side of this composition, where every sequence
of $N\OneL$ remains unchanged from the beginning up to the first
occurrence of an unambiguous class $c_u$.
Every following symbol is mapped to the empty string by means of
$[ ? \symp [\;] ] \;\kleenestar$ (eq. \ref{e-FSi}).
Finally, the extracted lower side is again converted into
2-level format\otherfootnote{n-formats}.

The extraction of the union $\lxu{\cup}S_m^e$ of extended middle
subsequences is performed in a similar way.

We then make the joint unions of initial
and extended middle subsequences\otherfootnote{n-regex}:
\begin{equation}
\lxu{\cup}S_i = \lxul{\cup}{s}S_i \; | \;
   [\; [\; \lxul{\cup}{n}S_i.u - \lxul{\cup}{s}S_i.u \;]
    \compose \lxul{\cup}{n}S_i \;]
 \label{e-ex-Si}
\end{equation}
\begin{equation}
\lxu{\cup}S_m^e = \lxul{\cup}{s}S_m^e \; | \;
   [\; [\; \lxul{\cup}{n}S_m^e.u - \lxul{\cup}{s}S_m^e.u \;]
    \compose \lxul{\cup}{n}S_m^e \;]
 \label{e-ex-Sme}
\end{equation}

\noindent
In both cases (eq. \ref{e-ex-Si} and \ref{e-ex-Sme})
we union all subsequences from the principal model $S$,
with all those subsequences from the auxiliary model $N$
that are not in $S$.

Finally, we generate the completed {\it s+n-type} transducer
from the joint unions of subsequences $\lxu{\cup}S_i$ and
$\lxu{\cup}S_m^e$, as decribed above
(eq. \ref{e-pre-sent-model}-\ref{e-fin-sent-model}).

A transducer completed in this way, disambiguates all subsequences known
to the principal incomplete s-type model, exactly as the underlying
HMM does, and all other subsequences as the auxiliary n-type model
does.

\vspace{3mm}
\section{An Implemented Finite-State Tagger \label{s-tagger}}
\vspace{1mm}

The implemented tagger requires three transducers
which represent a lexicon, a guesser and any above mentioned
approximation of an HMM.

All three transducers are sequential, i.e. deterministic on the
input side.

Both the lexicon and guesser unambiguously map a surface form
of any word that they accept to the corresponding class
of tags (fig.~\ref{f_tgrexm}, col.~1 and~2):
First, the word is looked for in the lexicon.
If this fails, it is looked for in the guesser.
If this equally fails, it gets the label {\tt [UNKNOWN]}
which associates the word with the tag class of unknown words.
Tag probabilities in this class
are approximated by tags of words that appear only once in the
training corpus.

As soon as an input token gets labelled with the tag class
of sentence end symbols (fig.~\ref{f_tgrexm}: {\tt [SENT]}),
the tagger stops reading words from the input.
At this point, the tagger has read and stored 
the words of a whole sentence
(fig.~\ref{f_tgrexm}, col.~1)
and generated the corresponding sequence of classes
(fig.~\ref{f_tgrexm}, col.~2).

The class sequence is now deterministically mapped
to a tag sequence (fig.~\ref{f_tgrexm}, col.~3)
by means of the HMM transducer.
The tagger outputs the stored word and tag sequence
of the sentence,
and continues in the same way with the remaining sentences
of the corpus.

\begin{figure}[htbp]
\begin{minipage}{80mm}
\begin{verbatim}
  The             [AT]            AT
  share           [NN,VB]         NN
  of              [IN]            IN
   ...             ...            ...
  tripled         [VBD,VBN]       VBD
  within          [IN,RB]         IN
  that            [CS,DT,WPS]     DT
  span            [NN,VB,VBD]     VBD
  of              [IN]            IN
  time            [NN,VB]         NN
  .               [SENT]          SENT
\end{verbatim}
\end{minipage}
\caption{Tagging a sentence \label{f_tgrexm}}
\end{figure}

\vspace{2mm}
\section{Experiments and Results \label{s-tests}}
\vspace{1mm}

This section compares different n-type and s-type transducers
with each other and with the underlying HMM.

The FSTs perform tagging faster than the HMMs.

Since all transducers are approximations of HMMs,
they give a lower tagging accuracy than the corresponding HMMs.
However, improvement in accuracy can be expected
since these transducers can be composed with transducers
encoding correction rules for frequent errors
(sec. \ref{s-intro}).

\begin{table*}[ht] \tabcolsep2mm
\begin{center}
\begin{math}
\begin{tabular}{|p{38mm}*{5}{|r}|} \cline{2-6}
  \multicolumn{1}{l|}{~} &
  accuracy &
  tagging speed &
  \multicolumn{2}{c|}{transducer size} &
  creation
\\ \cline{4-5}
  \multicolumn{1}{l|}{~} &
  in \% &
  in words/sec &
  \#~states &
  \#~arcs &
  time
\\ \hline \hline
HMM
  & 96.77 &  4~590 & \longbar & \longbar & \longbar \\ \hline \hline

n0-FST
  & 83.53 &{\bf 20~582}& 1 & 297 & 16 sec \\ \hline
n1-FST
  & 94.19 & 17~244 & 71 & 21~087 & 17 sec \\ \hline \hline

s+n1-FST~(20K,~F1)
  & 94.74 & 13~575 &   927 &   203~853 &  3 min \\ \hline
s+n1-FST~(50K,~F1)
  & 94.92 & 12~760 & 2~675 &   564~887 & 10 min \\ \hline
s+n1-FST~(100K,~F1)
  & 95.05 & 12~038 & 4~709 &   976~785 & 23 min \\ \hline

s+n1-FST~(100K,~F2)
  & 94.76 & 14~178 &   476 &   107~728 &  2 min \\ \hline
s+n1-FST~(100K,~F4)
  & 94.60 & 14~178 &   211 &    52~624 & 76 sec \\ \hline
s+n1-FST~(100K,~F8)
  & 94.49 & 13~870 &   154 &    41~598 & 62 sec \\ \hline
s+n1-FST~(1M,~F2)
  & 95.67 & 11~393 & 2~049 &   418~536 &  7 min  \\ \hline
s+n1-FST~(1M,~F4)
  & 95.36 & 11~193 &   799 &   167~952 &  4 min  \\ \hline
s+n1-FST~(1M,~F8)
  & 95.09 & 13~575 &   432 &    96~712 &  3 min  \\ \hline \hline
s+n1-FST ($\le 2$)
  & 95.06 & 8~180  &  9~796 & 1~311~962 & 39 min \\ \hline
s+n1-FST ($\le 3$)
  &{\bf 95.95}& 4~870 & 92~463 & 13~681~113 & 47 h \\ \hline
\end{tabular}
\end{math} \\

\small
\vspace{1mm}
\begin{math}
\begin{tabular}{|p{30mm} p{100mm}|} \hline
  Language:                 & English \\
  Corpora:                  & 19~944 words for HMM training,
                              19~934 words for test \\
  Tag set:                  & 74 tags~~ 297 classes \\ \hline
 \multicolumn{2}{|l|}{Types of FST (Finite-State Transducers)~: } \\
  ~~~~n0, n1                & n0-type (with only lexical probabilities)
                                or n1-type (sec. \ref{s-ntype}) \\
  ~~~~s+n1 (100K, F2)       & s-type (sec. \ref{s-stype}),
                               with subsequences of frequency $\ge$ 2,
                               from a training corpus of 100~000 words
                               (sec. \ref{s-make-stype}~a),
                               completed with n1-type (sec. \ref{s-compl-stype}) \\
  ~~~~s+n1 ($\le 2$)        & s-type (sec. \ref{s-stype}),
                               with all possible subsequences of length $\le$ 2 classes
                               (sec. \ref{s-make-stype}~b),
                               completed with n1-type (sec. \ref{s-compl-stype}) \\ \hline
  Computer:                & ultra2, 1 CPU, 512 MBytes physical RAM, 1.4 GBytes virtual RAM \\ \hline
\end{tabular}
\end{math} \\

\normalsize
\vspace{1mm}
\caption{Accuracy, speed, size and creation time of some HMM transducers
  \label{t_size}}
\end{center}
\end{table*}

Table \ref{t_size} compares different transducers on an English
test case.

The s+n1-type transducer containing all possible
subsequences up to a length of three classes is the most accurate
(table \ref{t_size}, last line, s+n1-FST ($\le 3$): 95.95~\%)
but also the largest one.
A similar rate of accuracy at a much lower size can be achieved with
the s+n1-type, either with all subsequences up to a length of two
classes (s+n1-FST ($\le 2$): 95.06~\%) or with subsequences
occurring at least once in a training corpus of 100~000 words
(s+n1-FST~(100K,~F1): 95.05~\%).

Increasing the size of the training corpus and the frequency limit,
i.e. the number of times that a subsequence must at least occur
in the training corpus in order to be selected
(sec. \ref{s-make-stype}~a), improves the relation between tagging
accuracy and the size of the transducer.
E.g. the s+n1-type transducer that encodes subsequences from a
training corpus of 20~000 words
(table \ref{t_size}, s+n1-FST~(20K,~F1): 94.74~\%, 927 states, 203~853 arcs),
performs less accurate tagging and is bigger
than the transducer that encodes subsequences occurring at least eight
times in a corpus of 1~000~000 words
(table \ref{t_size}, s+n1-FST~(1M,~F8): 95.09~\%, 432 states, 96~712 arcs).

Most transducers in table \ref{t_size} are faster then the
underlying HMM; the n0-type transducer about five times\footnote{
Since n0-type and n1-type transducers have deterministic
states only, a particular fast matching algorithm can be
used for them.}.
There is a large variation in speed between the different
transducers due to their structure and size. 

\begin{table*}[ht] \tabcolsep2mm
\begin{center}
\begin{math}
\begin{tabular}{|p{40mm}*{6}{|r}|} \cline{2-7}
\multicolumn{1}{l|}{~} &
  \multicolumn{6}{c|}{accuracy in \%} \\ \cline{2-7}
\multicolumn{1}{l|}{~} &
   English & Dutch  & French & German & Portug. & Spanish \\ \hline \hline

HMM
  & 96.77 & 94.76 & 98.65 & 97.62 & 97.12 & 97.60 \\ \hline \hline

n0-FST
  & 83.53 & 81.99 & 91.13 & 82.97 & 91.03 & 93.65 \\ \hline
n1-FST
  & 94.19 & 91.58 & 98.18 & 94.49 & 96.19 & 96.46 \\ \hline

s+n1-FST~(20K,~F1)
  & 94.74 & 92.17 & 98.35 & 95.23 & 96.33 & 96.71 \\ \hline
s+n1-FST~(50K,~F1)
  & 94.92 & 92.24 &{\bf 98.37}& 95.57 & 96.49 & 96.76 \\ \hline
s+n1-FST~(100K,~F1)
  &{\bf 95.05}&{\bf 92.36}&{\bf 98.37}& 95.81 &{\bf 96.56}&{\bf 96.87} \\ \hline

s+n1-FST~(100K,~F2)
  & 94.76 & 92.17 & 98.34 & 95.51 & 96.42 & 96.74 \\ \hline
s+n1-FST~(100K,~F4)
  & 94.60 & 92.02 & 98.30 & 95.29 & 96.27 & 96.64 \\ \hline
s+n1-FST~(100K,~F8)
  & 94.49 & 91.84 & 98.32 & 95.02 & 96.23 & 96.54 \\ \hline

s+n1-FST~($\le 2$)
  &{\bf 95.06}&  92.25 &{\bf 98.37}&{\bf 95.92}&  96.50 &{\bf 96.90} \\ \hline \hline

HMM train.crp.~(\#wd)
  & 19~944 & 26~386 & 22~622 & 91~060 & 20~956 & 16~221 \\ \hline
test corpus~(\# words)
  & 19~934 & 10~468 &  6~368 & 39~560 & 15~536 & 15~443 \\ \hline
\# tags
  &     74 &     47 &     45 &     66 &     67 &     55 \\ \hline
\# classes
  &    297 &    230 &    287 &    389 &    303 &    254 \\ \hline
\end{tabular}
\end{math} \\

\small
\vspace{1mm}
\begin{math}
\begin{tabular}{|p{80mm} p{45mm}|} \hline
 Types of FST (Finite-State Transducers)~:  & cf. table \ref{t_size} \\ \hline
\end{tabular}
\end{math} \\

\normalsize
\vspace{1mm}
\caption{Accuracy of some HMM transducers for different languages
  \label{t_aclang}}
\end{center}
\end{table*}

Table \ref{t_aclang} compares the tagging accuracy of different
transducers and the underlying HMM for different languages.
In these tests the highest accuracy was always obtained by
s-type transducers, either with all subsequences up to
a length of two classes\footnote{
A maximal length of three classes is not considered here
because of the high increase in size and a low increase
in accuracy.}
or with subsequences occurring at least once in a corpus of
100~000 words.

\vspace{3mm}
\section{Conclusion and Future Research \label{s-concl}}
\vspace{1mm}

The two methods described in this paper allow the approximation
of an HMM used for part-of-speech tagging, by a finite-state transducer.
Both methods have been fully implemented.

The tagging speed of the transducers is up to five times higher
than that of the underlying HMM.

The main advantage of  transforming an HMM is that the resulting
FST can be handled by finite state calculus\footnote{
A large library of finite-state functions is available at Xerox.}
and thus be directly composed with other transducers which
encode tag correction rules and/or perform further steps of
text analysis.

{\bf Future research} will mainly focus on this possibility
and will include composition with, among others:

\begin{itemize}
\item
Transducers that encode correction rules
(possibly including long-distance dependencies)
for the most frequent tagging errors,
in order to significantly improve tagging accuracy.
These rules can be either
extracted automatically from a corpus (Brill, 1992)
or written manually (Chanod and Tapanainen, 1995).
\item
Transducers for light parsing, phrase extraction and other analysis
(A\"{\i}t-Mokhtar and Chanod, 1997).
\end{itemize}

An HMM transducer can be composed with one or more of these transducers
in order to perform complex text analysis using only a single transducer.

We also hope to improve the n-type model by using look-ahead
to the following tags\footnote{
Ongoing work has shown that, looking ahead to just one tag
is worthless because it makes tagging results highly ambiguous.}.

\newpage
\section*{Acknowledgements}

I wish to thank the anonymous reviewers of my paper for their valuable
comments and suggestions.

I am grateful to Lauri Karttunen and Gregory Grefenstette (both
RXRC Grenoble) for extensive and frequent discussion
during the period of my work,
as well as to Julian Kupiec (Xerox PARC) and Mehryar Mohri
(AT\&T Research) for sending me some interesting ideas
before I started.

Many thanks to all my colleagues at RXRC Grenoble who helped me
in whatever respect, particularly to Anne Schiller, Marc Dymetman
and Jean-Pierre Chanod for discussing parts of the work,
and to Irene Maxwell for correcting various versions of the paper.

\newpage
\section*{References}
\vspace{1.5mm}

\mybibbegin

\mybibitem
A\"{\i}t-Mokhtar, Salah and Chanod, Jean-Pierre (1997).
Incremental Finite-State Parsing.
In the {\it Proceedings of the 5th Conference of
Applied Natural Language Processing.}\/
ACL, pp. 72-79.
Washington, DC, USA.

\mybibitem
Bahl, Lalit R. and Mercer, Robert L. (1976).
Part of Speech Assignment by a Statistical Decision Algorithm.
In {\it IEEE international Symposium on Information Theory}\/.
pp. 88-89. Ronneby.

\mybibitem
Brill, Eric (1992).
A Simple Rule-Based Part-of-Speech Tagger.
In the {\it Proceedings of the 3rd conference on Applied Natural
Language Processing}\/, pp. 152-155.
Trento, Italy.

\mybibitem
Chanod, Jean-Pierre and Tapanainen, Pasi (1995).
Tagging French - Comparing a Statistical and a Constraint Based Method.
In the {\it Proceedings of the 7th conference of the EACL}\/, pp. 149-156.
ACL. Dublin, Ireland.

\mybibitem
Church, Kenneth W. (1988).
A Stochastic Parts Program and Noun Phrase Parser for
Unrestricted Text.
In {\it Proceedings of the 2nd Conference on Applied Natural
Language Processing}\/. ACL, pp. 136-143.

\mybibitem
Kaplan, Ronald M. and Kay, Martin (1994). 
Regular Models of Phonological Rule Systems.
In {\it Computational Linguistics}\/.
20:3, pp. 331-378.

\mybibitem
Karttunen, Lauri (1995).
The Replace Operator.
In the {\it Proceedings of the 33rd Annual Meeting of the
Association for Computational Linguistics}\/.
Cambridge, MA, USA.
{\tt cmp-lg/9504032}

\mybibitem
Kempe, Andr\'e and Karttunen, Lauri (1996).
Parallel Replacement in Finite State Calculus.
In the {\it Proceedings of the 16th International Conference
on Computational Linguistics}\/, pp. 622-627.
Copenhagen, Denmark.
{\tt cmp-lg/9607007}

\mybibitem
Rabiner, Lawrence R. (1990).
A Tutorial on Hidden Markov Models and Selected Applications
in Speech Recognition.
In {\it Readings in Speech Recognition}\/ (eds. A. Waibel, K.F. Lee).
Morgan Kaufmann Publishers, Inc. San Mateo, CA., USA.

\mybibitem
Roche, Emmanuel and Schabes, Yves (1995).
Deterministic Part-of-Speech Tagging with Finite-State Transducers.
In {\it Computational Linguistics}\/. Vol. 21, No. 2, pp. 227-253.

\mybibitem
Viterbi, A.J. (1967).
Error Bounds for Convolutional Codes and an Asymptotical Optimal
Decoding Algorithm.
In {\it Proceedings of IEEE}\/, vol. 61, pp. 268-278.

\mybibend

\newpage
\section*{{\sc Annex:} Regular Expression Operators}
\vspace{1.7mm}

Below, {\tt a} and {\tt b} designate symbols,
{\tt A} and {\tt B} designate languages, and
{\tt R} and {\tt Q} designate relations between two languages.
More details on the following operators and pointers to
finite-state literature can be found in
{\tt http://www.rxrc.xerox.com/research/mltt/fst}

\begin{table}[htbp] \tabcolsep0mm
\begin{center}
\begin{math}
\begin{tabular}{p{15mm}p{64mm}}
{\tt \contains A}  &
  Contains.
    Set of strings containing at least one occurrence of
    a string from {\tt A} as a substring.  \\
{\tt $\xnot$A}  &
  Complement (negation).
    All strings except those from {\tt A}.  \\
{\tt $\bs$a}  &
  Term complement.
    Any symbol other than {\tt a}.  \\
{\tt A*}  &
  Kleene star.
    Zero or more times {\tt A} concatenated with itself.  \\
{\tt A+}  &
  Kleene plus.
    One or more times {\tt A} concatenated with itself.  \\
{\tt a{\SmSpc}->{\SmSpc}b}  &
  Replace.
    Relation where every {\tt a} on the upper side gets mapped
    to a {\tt b} on the lower side. \\
{\tt a{\SmSpc}<-{\SmSpc}b}  &
  Inverse replace.
    Relation where every {\tt b} on the lower side gets mapped
    to an {\tt a} on the upper side. \\
{\tt a:b}  &
  Symbol pair
    with {\tt a} on the upper and {\tt b} on the lower side.  \\
{\tt $\OLPair{a}{b}$}  &
  1-Level symbol
    which is the 1-level form ($\OneL$) of the symbol pair {\tt a:b}.  \\
{\tt R.u} &
  Upper language
    of {\tt R}. \\
{\tt R.l} &
  Lower language
    of {\tt R}. \\
{\tt A~~B} &
  Concatenation
    of all strings of {\tt A} with all strings of {\tt B}. \\
{\tt A | B} &
  Union
    of {\tt A} and {\tt B}. \\
{\tt A \& B} &
  Intersection
    of {\tt A} and {\tt B}. \\
{\tt A - B} &
  Relative complement (minus).
    All strings of {\tt A} that are not in {\tt B}. \\
{\tt A{\SmSpc}.x.{\SmSpc}B} &
  Cross Product (Cartesian product)
    of the languages {\tt A} and {\tt B}. \\
{\tt R{\SmSpc}.o.{\SmSpc}Q} &
  Composition
    of the relations {\tt R} and {\tt Q}. \\
{\tt R$\OneL$}  &
  1-Level form.
    Makes a language out of the relation {\tt R}.
    Every symbol pair becomes a simple symbol.
    (e.g. {\tt a:b} becomes {\tt $\OLPair{a}{b}$} and
     {\tt a} which means {\tt a:a} becomes {\tt $\OLPair{a}{a}$}) \\
{\tt A$\TwoL$}  &
  2-Level form.
    Inverse operation to $\OneL$ ~({\tt R$\OneL\TwoL=\;$R}).  \\
\mbox{\tt 0 {\it or} [~]}  &
  Empty string (epsilon).  \\
{\tt ?}  &
  Any symbol
    in the known alphabet and its extensions \\
\end{tabular}
\end{math}
\end{center}
\end{table}

\end{document}